\documentclass{article}
\usepackage{spconf,amsmath,graphicx}
\usepackage{amssymb}
\usepackage{multirow}
\usepackage{booktabs}
\usepackage{url}

\newcommand{\STAB}[1]{\begin{tabular}{@{}c@{}}#1\end{tabular}}

\title{Integrated Grad-CAM: Sensitivity-Aware Visual Explanation of Deep Convolutional Networks via Integrated Gradient-Based Scoring}
\name{%
\begin{tabular}{@{}c@{}}
Sam Sattarzadeh$^{\star}$, 
Mahesh Sudhakar$^{\star}$,
Konstantinos N. Plataniotis$^{\star}$,
\\
Jongseong Jang$^{\dagger}$, Yeonjeong Jeong$^{\dagger}$, Hyunwoo Kim$^{\dagger}$
\end{tabular}}  
  
\address{$^{\star}$Department of Electrical \& Computer Engineering, University of Toronto\\ $^{\dagger}$Fundamental Research Lab, LG AI Research}

\begin{document}
%
\maketitle
\begin{abstract}
Visualizing the features captured by Convolutional Neural Networks (CNNs) is one of the conventional approaches to interpret the predictions made by these models in numerous image recognition applications. Grad-CAM is a popular solution that provides such a visualization by combining the activation maps obtained from the model. However, the average gradient-based terms deployed in this method underestimates the contribution of the representations discovered by the model to its predictions. Addressing this problem, we introduce a solution to tackle this issue by computing the path integral of the gradient-based terms in Grad-CAM. We conduct a thorough analysis to demonstrate the improvement achieved by our method in measuring the importance of the extracted representations for the CNN's predictions, which yields to our method's administration in object localization and model interpretation.
\end{abstract}

\section{Introduction}
\label{sec:intro}

Despite the strong ability of Convolutional Neural Networks (CNNs) in feature representation and image recognition, these cumbersome models often lack explainability, limiting the trust and reliance of the end-users towards the decisions made by them. Explainable AI (XAI) is a field that attempts to make the third-party consumers trusted on AI models by opening their black-box and elucidating the reasoning of the models for their predictions. By meeting these goals, XAI algorithms provide the users with an answer to questions such as ``Why does the model predict what it predicts?", ``When does the model make an unreliable prediction?", ``How does the model behave if it is put in a specific scenario?" etc. \cite{lipton2018mythos, guidotti2018survey}.

In particular, visual explanation methods (a.k.a. attribution methods) are among the most celebrated groups of XAI methods that explain the predictions made by CNNs. These algorithms are a branch of `post-hoc' explanation algorithms that interpret the behavior of the model in the evaluation phase. Visual explanation methods formulate their problem as follows: They take a model trained for image recognition and a digital image as inputs. The model is fed with the image and makes a prediction accordingly. The method's objective is to output a 2-dimensional heatmap named `explanation map' with the same height and width as the input image. The explanation map valuates the regions of the image, based on their contribution in the model's prediction.

One notable group of visual XAI approaches are the ones based on the Class Activation Mapping (CAM) method \cite{CAM}. These approaches are specialized for CNNs and inspired from \cite{zhou2014object} which showed that CNNs act like object detectors and can learn high-level representations of the object instances in an unsupervised manner. Grad-CAM is a popular CAM-based approach that utilizes backpropagation to score the feature maps' locations in a specific layer \cite{GradCAM}.
Grad-CAM and the other methods employing backpropagation to form explanation maps (such as Grad-CAM++ and XGrad-CAM \cite{GradCAM, fu2020axiom}), offer great versatility and faithfulness. However, the performance of these methods is limited as gradient-based values underestimate the sensitivity of the model's output to the features represented in the image. This shortcoming has been addressed in prior works such as \cite{ScoreCAM, AblationCAM, SmoothGradCAMPP}.

\begin{figure}[t]
    \centering
    \includegraphics[width=0.9\linewidth]{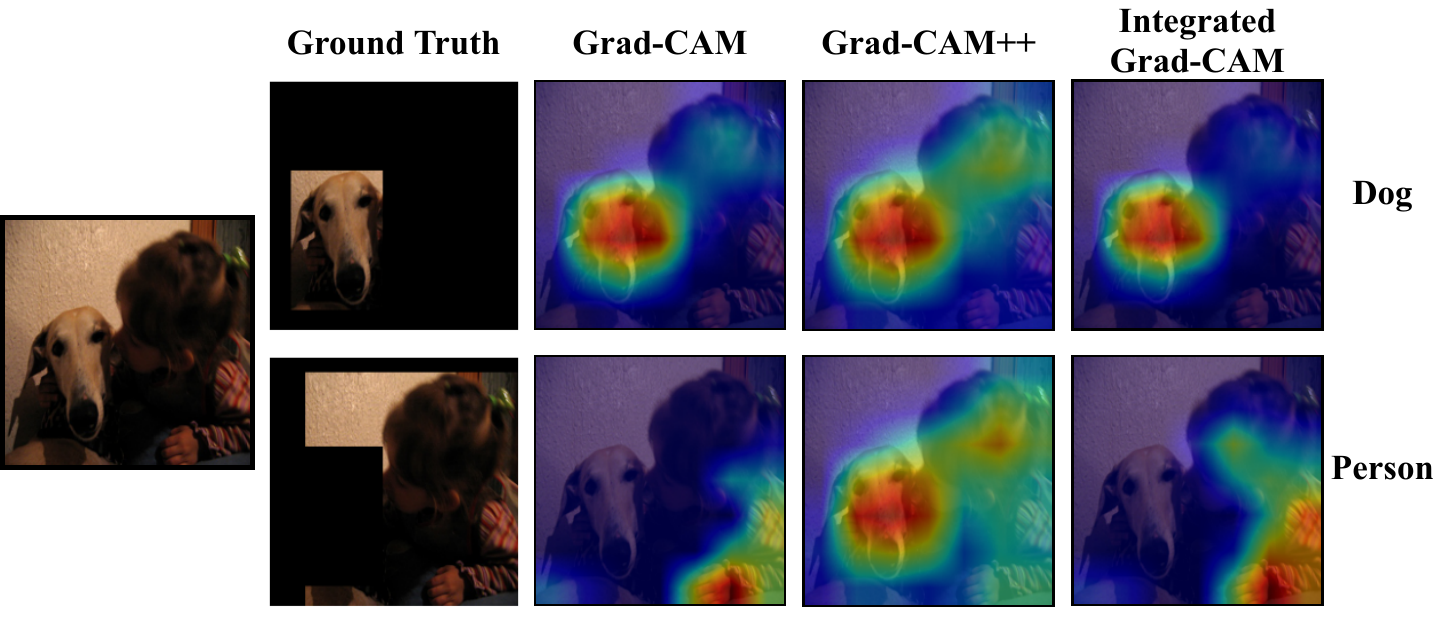}
    \caption{Comparison of baseline CAM-based methods with Integrated Grad-CAM to show the ability of our method to generate faithful class discriminative explanation maps.}
    \label{fig:my_label}
\end{figure}
 
In this work, we propose a novel technique to reduce the shortcomings of Grad-CAM. In common with Grad-CAM and Grad-CAM++, our method also utilizes signal backpropagation for weighting feature maps. However, we replace the gradient terms in Grad-CAM with similar terms based on \textit{Integrated Gradient}, inspired by an attribution method of the same name \cite{IntegGrad}. Hence, we name our CAM-based algorithm \textit{Integrated Grad-CAM}. To summarize, the main contributions of this work are as follows:
\begin{itemize}
    \item We propose Integrated Grad-CAM, which bridges Integrated Gradient and Grad-CAM to solve the gradient issues in the prior CAM-based methods taking benefits of backpropagation techniques.
    \item We demonstrate our proposed method's ability, compared to Grad-CAM, Grad-CAM++, and Integrated Gradient, by conducting experiments on shallow and deep networks and performing qualitative and quantitative metrics. We achieve the empirical results implying that our method successfully combines the practical ideas in each of these methods to improve them in completeness, faithfulness, and satisfaction.
\end{itemize}

\section{Related works}
\label{sec:format}

\subsection{Backpropagation-based methods: } Computing the gradient of a model’s output to the input features or the hidden neurons is the basis of this type of algorithms. The earliest backpropagation-based methods operate directly by computing the sensitivity of the model’s confidence score to the input features \cite{simonyan2013deep}. To develop such methods, some approaches such as \cite{bach2015pixel, nam2020relative} modify their backpropagation rules to assign scores to the input features denoting the relevance or irrelevance of the input features to the model’s prediction. Also, an Integrated Gradient calculation was defined by \cite{IntegGrad}, to satisfy to axioms termed as \textit{sensitivity} and \textit{implementation invariance} as per their definition.

\subsection{Grad-CAM: } This method runs in two steps to form an explanation map using the outputs of a given layer (usually, the last convolutional layer) of the target CNN model. in the feature extraction unit of the model. In the first step, the selected layer is probed, and their corresponding feature maps are collected. In the second step, the signal is partially backpropagated from the output to the selected layer. Then, the average of the gradient values with respect to the pixels in each feature maps are calculated. Assume the input image to be $I$, and the class confidence score of the model for class $c$ to be $y_{c}(I)$, and a layer $l$ selected, Grad-CAM initially collects the feature maps $\{A^{l1}(I), A^{l2}(I),..., A^{lN}(I)\}$ in a forward pass ($N$ denotes the number of feature maps in the chosen layer). Then, the signal is passed back from the output neuron to the layer $l$. To reach the explanation map, Grad-CAM performs a weighted combination of the feature maps using their corresponding average gradient-based weights:
\begin{equation}
    M_{Grad-CAM}^c = \text{ReLU}\big( \sum_{k=1}^{N} (\frac{1}{Z}\sum_{i,j}\frac{\partial y_c(I)}{\partial A_{ij}^{lk}(I)})A^{lk}(I)\big)
    \label{grad-cam}
\end{equation}
In the equation above, $A_{ij}^{lk}(I)$ refers to the location $\{i,j\}\in \mathbb{R}^{u,v}$ in the $k$-th feature map and $\{u,v\}$ denote the dimensions of the feature maps ($Z=u\times v$). The dimensions of $M_{Grad-CAM}^c$ is the same as that of the feature maps, and usually smaller than the input image. Hence, the final Grad-CAM explanation map is reached by upsampling $M_{Grad-CAM}^c$ to the size of $I$ through bilinear interpolation.

\subsection{Integrated Gradient}
One of the main drawbacks of deploying backpropagation in attribution methods is that they violate \textit{sensitivity} axiom. As discussed in previous works such as Integrated Gradient and DeepLift \cite{IntegGrad, shrikumar2016not}, this axiom implies that for each given pair of input and baseline image differing only in one feature, an attribution method should highlight this difference by assigning different values corresponding to that feature, which envisions the response of the model to this difference. To address this issue in vanilla gradient \cite{simonyan2013deep}, it was proposed in \cite{IntegGrad} that given a defined baseline, and the input image, the sensitivity of output's confidence scores to input features can be justified stronger by calculating the integral of gradient values on any continuous path connecting the baseline and the input.

\section{Methodology}
\label{sec:pagestyle}
The same as gradient-based methods, Grad-CAM breaks the sensitivity axiom \cite{IntegGrad} while dealing with non-linear components of a CNN, such as activation functions (e.g., ReLU). To reduce this problem, we integrate the local sensitivity scores of the model's output to the neurons in each feature map when the input image is scaled from a pre-defined baseline $I'$ to the main input image $I$. Given a pair of baseline and input, a path connecting these two is defined as:
\begin{equation}
    \gamma(\alpha)=I'+f(\alpha)\times(I-I')
    \label{eq:path}
\end{equation}
where $\alpha$ is a scalar variable, and the function $f(\alpha) : \mathbb{R}\rightarrow\mathbb{R}$ is differentiable and monotonically increasing when $0\leq\alpha\leq 1$, and satisfies $f(0)=0$ and $f(1)=1$. Gradient-based schemes may fail to quantify neurons' contribution in predicting an output for $I$ correctly when some of the paths linking them with the output node possess inactivated neurons \cite{shrikumar2016not}. Hence, the neurons' contribution scores can be determined more accurately via probing the relationship between them and the output node when the input image changes from a certain baseline. For each pair of assumed functions $g(.)$ and $h(.)$, path integral gradient (PathIG) are calculated as follows:
\begin{equation}
    \text{PathIG}_{h,g}(I)\equiv \int_{\alpha=0}^{1}\frac{d h(\gamma(\alpha))}{d g(\gamma(\alpha))}[g(\gamma(\alpha))-g(I')]d\alpha
    \label{eq:PathIG}
\end{equation}
For more simplicity, the path from the baseline to the input image is defined as a straight linear path for computation simplicity by setting $f(\alpha)=\alpha$.
In Integrated Grad-CAM, we formulate the scoring scheme considering average gradient values for each feature map. The general formulation of our equation is similar to eq. \eqref{grad-cam}. However, we update the average gradient terms in Grad-CAM with corresponding average integrated gradient values. We consider a straight linear path in the image domain from the reference image $I'$ to the desired input image to simplify our formulation. Hence, our explanation maps $M^c$ are computed as:
\begin{equation}
    M^c=\int_{\alpha=0}^{1}\text{ReLU}(\sum_{k=1}^{N}\sum_{i,j}\frac{\partial y_c(\gamma(\alpha))}{\partial A_{ij}^{lk}(\gamma(\alpha))}\Delta_{lk}(\gamma(\alpha)))d\alpha
    \label{int-grad-cam}
\end{equation}
where,
\begin{equation}
    \Delta_{lk}(\gamma(\alpha))=(A^{lk}(\gamma(\alpha))-A^{lk}(I'))
    \label{delta}
\end{equation}

In the equations above, $y_{c}(\gamma(\alpha))$ is the confidence score achieved for class $c$ and the input image $\gamma(\alpha)$, and $A^{lk}(\gamma(\alpha))$ is the $k$-th feature map derived from the layer $l$. Also, according to \cite{IntegGrad}, a black image is an appropriate choice for a baseline since black regions contain no significant attributions. The same as Grad-CAM, as far as our saliency maps are generated throughout the equation above, our explanation maps are reached after upsampling $M^c$ to the dimensions of the input image, via bilinear interpolation.

\begin{figure}[t]
    \centering
    \includegraphics[width=0.99\linewidth]{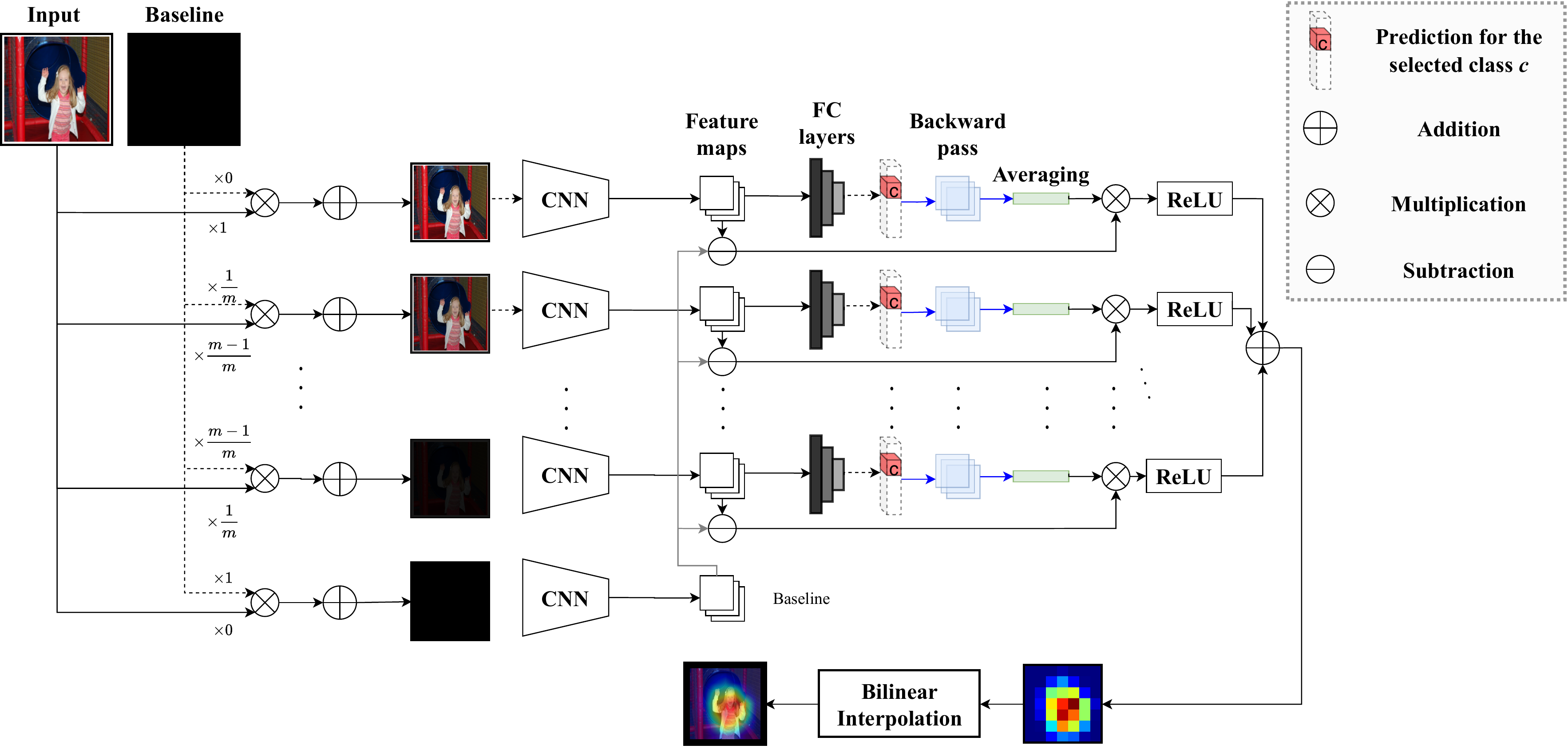}
    \caption{Schematic of the proposed method considering that the baseline image is set to black and the path connecting the baseline and the input is set as a straight line.}
    \label{fig:Schematic}
\end{figure}

Implementing integral functions on a software (or hardware) environment has always been a challenging task. In our case, a simple solution to overcome this issue is to approximate the integral in equation \eqref{int-grad-cam} with a summation via Riemman approximation. To perform such an estimation, we sample points along the path with a constant interval, calculate the expression in equation \eqref{int-grad-cam} for these points, and estimate the term $d\alpha$ with the interval size. Considering the interval step to be $\frac{1}{m} (m \in \mathbb{N})$, the integrated gradient-based score maps can be approximated as follows:
\begin{equation}
    M^c \approx\sum_{t=1}^{m}\text{ReLU}\big(\frac{1}{m}\sum_{k=1}^{N}\sum_{i,j}\frac{\partial y_c(\gamma(\frac{t}{m}))}{\partial A_{ij}^{lk}(\gamma(\frac{t}{m}))})\Delta(\gamma(\frac{t}{m}))\big)
    \label{int-grad-cam-approx}
\end{equation}

Solving the equation \eqref{int-grad-cam} using the equation above makes our method equivalent to averaging Grad-CAM saliency maps reached for multiple copies of the input, which are linearly interpolated with the defined baseline, as shown in figure \ref{fig:Schematic}.

\section{Experiments}
\label{sec:typestyle}
To verify the improved completeness and faithfulness of the explanations provided by our method, we have conducted experiments that compare our method with the baseline methods, Grad-CAM, and Grad-CAM++. In the experiments, we utilized TorchRay library provided in \cite{fong2019understanding}, and implemented our method in PyTorch \cite{paszke2019pytorch}\footnote{Our code is publically available at: \url{https://github.com/smstrzd/IntegratedGradCAM}}. In all experiments, we applied our method and other conventional CAM-based algorithms. We selected the last convolutional layer since this layer provides the highest-level representations captured by CNN. Moreover, we set the interval step $m$ in our method to 50 to reach an acceptable trade-off between precision and computational overhead. However, in the case that this parameter is set to any number between 20 and 200, the results of applying our method do not vary considerably.

\subsection{Dataset and Models}
\label{ssec:subhead}
Our experiments are performed on two networks trained on PASCAL VOC 2007 dataset. We used the test set of this database to collect the qualitative and quantitative results. PASCAL VOC 2007 is an object detection dataset, containing 4952 test images from 20 different output classes. The presence of multiple objects from either the same instance or different instances makes interpreting the models trained on this dataset more challenging so that the explanation approaches producing class-indiscriminative saliency maps for model's prediction for multiple classes are expected to fail to interpret the models trained on this dataset accurately.

In this work, we utilized two networks with different structures, trained on the mentioned dataset by \cite{ModelTrainer} and provided in TorchRay library. The first model is a VGG-16 network achieving a top-1 accuracy of 87.18\%, and the latter model is a deeper ResNet-50 network with a top-1 accuracy of 87.96\%. Both models take images of size $224\times 224\times 3$ as input. Thus, all images are resized to these dimensions before they are passed through the models.

\subsection{Quantitative Evaluation}
\begin{table}[t]
 \centering
 \begin{tabular}{c c c c c}
 \toprule
  & \multirow{2}{*}{\textbf{Metric}} &
 Grad- & Grad- & Integrated \\
 & &  CAM & CAM++ & Grad-CAM  \\
 \midrule
\multirow{4}{*}{\STAB{\rotatebox[origin=c]{90}{\textbf{VGG16}}}} &
 \textbf{EBPG} &
 55.44 & 46.29 & \textbf{55.94} \\
  & \textbf{Bbox} &
 51.7 & 55.59 & \textbf{55.6} \\
 & \textbf{Drop\%} & 49.47 & 60.63 & \textbf{47.96} \\
  & \textbf{Increase\%} & 31.08 & 23.89 &  \textbf{31.47} \\
 \midrule
\multirow{4}{*}{\STAB{\rotatebox[origin=c]{90}{\textbf{ResNet-50}}}} &
 \textbf{EBPG} &
60.08 & 47.78 & \textbf{60.41} \\
  & \textbf{Bbox} &
 60.25 & 58.66 & \textbf{61.94} \\
  & \textbf{Drop\%} &
 35.80 & 41.77 & \textbf{34.49} \\
  & \textbf{Increase\%} &
 36.58 & 32.15 & \textbf{36.84} \\
 \bottomrule
 \end{tabular}
 \caption{Results of quantitative analysis on PASCAL VOC 2007 test set. For each metric, the best is shown in bold. Except for Drop\%, the higher is better for all other metrics. The results are reported in percentage.}
 \label{tab: gt_metrics}
\end{table}
\label{sssec:subsubhead}
To compare our method with the other state-of-the-art CAM-based methods, we utilize two types of quantitative metrics. First, we deploy ground truth-based metrics, including Energy-based Pointing game (\textbf{EBPG}) and Bounding box (\textbf{Bbox}), to assess our method's ability in accurate object localization and feature visualization, compared to the baseline methods. Besides, we measure ``\textbf{Drop\%}" and ``\textbf{Increase\%}" to evaluate the faithfulness of the explanations by observing the model's behavior when it is fed only with the features denoted as important by an explanation algorithm. The description of the metrics is provided below.
\subsubsection {Ground truth-based metrics}
Energy-based pointing game which is developed by \cite{ScoreCAM}, quantifies the fraction of energy in each resultant explanation map $S$  captured in the corresponding ground truth mask $G$, as $EBPG = \frac{||S \odot G||_{1}}{||S||_{1}}$.
On the other hand, Bounding box, as introduced by \cite{Schulz2020Restricting} is a size-adaptive variant of mIoU. Denoting $N$ as the number of ground truth pixels in $G$, Bbox score is calculated by counting the fraction of pixels in $S$ among the highest $N$ pixels which are located inside the mask $G$.
\subsubsection{Drop/Increase rate}
As introduced in \cite{GradCAMPP} and developed by \cite{AblationCAM}, these metrics measure the correlation of the explanation maps generated by explanation algorithms with the model's prediction scores, by quantifying the positive attributions captured and the negative attribution discarded, respectively. Given a model $\Psi(.)$, an input image $I_{i}$ from a dataset containing $K$ images, and an explanation map $S(I_{i})$, initially a threshold function $T(.)$ is applied on $S(I_{i})$ to extract the most important 15\% pixels (based on $S(I_{i})$) from $I_{i}$ using point-wise multiplication. The confidence scores on the masked images are then compared with the original scores as follows:
\begin{equation}
    Drop\%=\frac{100}{K}\sum_{i=1}^{K}\frac{\text{ReLU}(\Psi(I_{i})-\Psi(I_{i}\odot T(I_{i})))}{\Psi(I_{i})}
\end{equation}
\begin{equation}
    Increase\%=\frac{100}{K}\sum_{i=1}^{K} \text{sign}(\Psi(I_{i}\odot T(I_{i}))-\Psi(I_{i}))
\end{equation}

\subsection{Discussion}
\begin{figure}[t]
    \centering
    \includegraphics[width=0.8\linewidth]{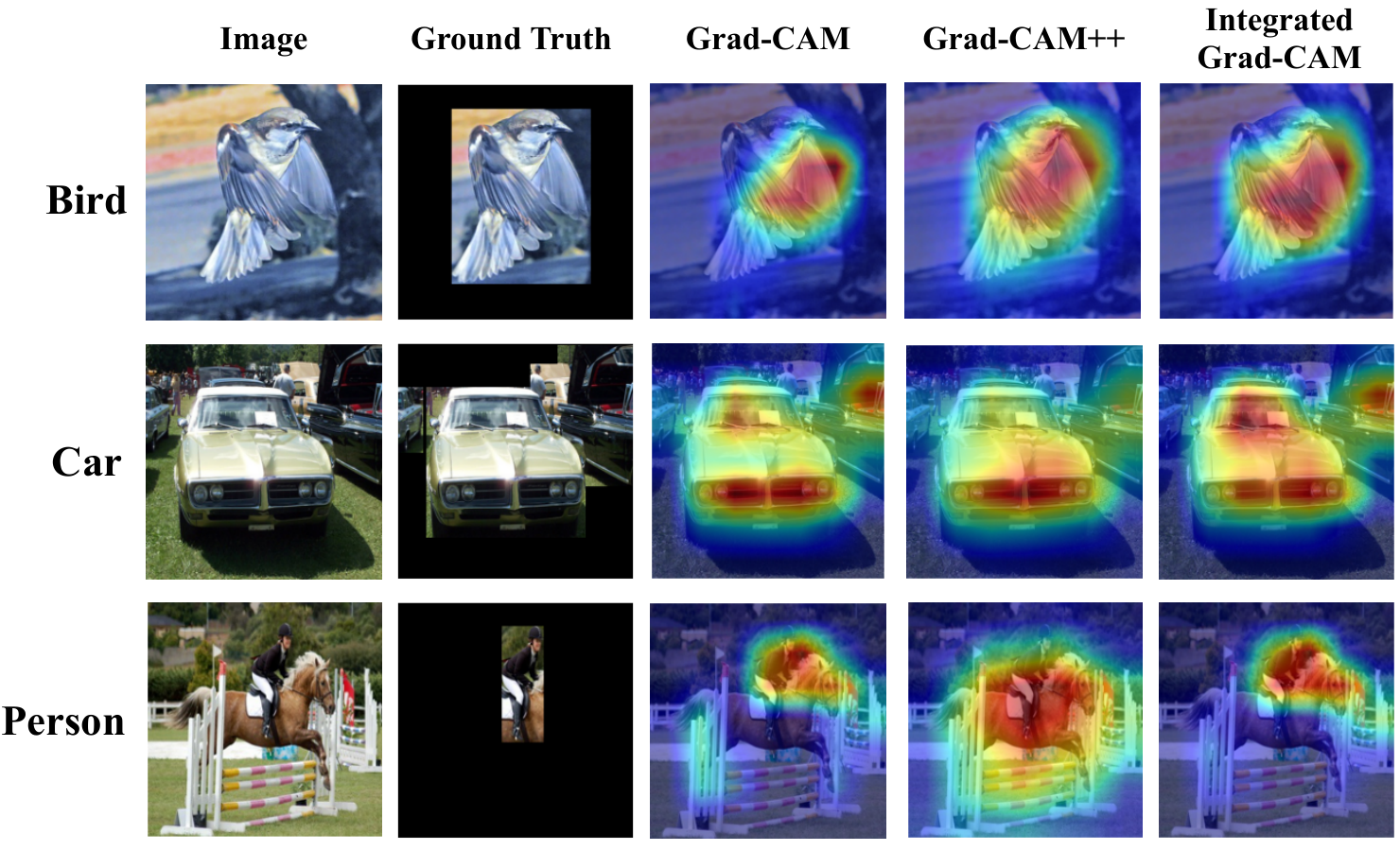}
    \caption{Qualitative comparison of baseline CAM-based XAI methods with Integrated Grad-CAM (our proposed). The sample images are given to a ResNet-50 model trained on PASCAL VOC 2007 dataset \cite{PASCALVOC}.}
    \label{fig:QRS}
\end{figure}

Every concrete explanation should satisfy two properties that are ``faithfulness" and ``understandability". Faithfulness denotes that explanations should reflect the exact behavior of the target model, while understandability means that explanations should be interpretable enough from the users' end. Our developed method is able to satisfy faithfulness and understandability better than Grad-CAM and Grad-CAM++. This is verified in table \ref{tab: gt_metrics} by model truth-based and ground truth-based metrics, respectively. Also, as shown in Figs. \ref{fig:my_label} and \ref{fig:QRS}, our method has a greater ability in highlighting more crucial attributions, compared to the conventional methods. The qualitative images are for the ResNet-50 model, though our method's advantages are also visible on the VGG-16 model.


Despite of its superior performance, our method provides more computational overhead compared rather than Grad-CAM and Grad-CAM++. Conducting a complexity evaluation on 100 random images from PASCAL VOC 2007 test set given to the ResNet-50 model, it was observed that both of these methods run in 11.3 milliseconds on a P100-PCIe GPU with 16GB of memory, while Integrated Grad-CAM (with its interval step set to $20$) requires 54.8 milliseconds in average to operate on each image. Increasing the interval step will slow down the method more, without any significant change in the reached explanation maps.


\section{Conclusion}
\label{sec:illust}
To deal with the fact that gradient-based CNN visualization approaches such as Grad-CAM are prone to miscalculate the features' value, we proposed Integrated Grad-CAM. Our method showed the ability to correct the measurements for scoring the attributions captured by CNN since it applies the path integral of a defined gradient-based term. Our experiments show that our approach improves Grad-CAM both in precise localization of the object regions and interpreting the predictions made by CNNs.


\bibliographystyle{IEEEbib}
\bibliography{refs}
\end{document}